\documentclass[twoside,11pt]{article}

\usepackage{jmlr2e}
\usepackage{graphicx}
\usepackage{wasysym}
\usepackage{url}

\ShortHeadings{Proposition of a Theoretical Model for Missing Data Imputation}{Leke, Paul and Marwala}
\firstpageno{1}

\begin{document}

\title{Proposition of a Theoretical Model for Missing Data Imputation using Deep Learning and Evolutionary Algorithms}

\author{\name Collins Leke \email collinsl@uj.ac.za \\
	\name Tshilidzi Marwala \email tmarwala@uj.ac.za \\
	\addr Department of Electrical and Electronic Engineering Science\\
	University of Johannesburg\\
	Johannesburg, PO Box 524, Auckland Park, 2006, South Africa
	\AND
	\name Satyakama Paul \email psatyakama@student.uj.ac.za \\
	\addr Department of Mechanical Engineering Science\\
	University of Johannesburg\\
	Johannesburg, PO Box 524, Auckland Park, 2006, South Africa
}


\maketitle

\begin{abstract}
In the last couple of decades, there has been major advancements in the domain of missing data imputation. The techniques in the domain include amongst others: Expectation Maximization, Neural Networks with Evolutionary Algorithms or optimization techniques and K-Nearest Neighbor approaches to solve the problem. The presence of missing data entries in databases render the tasks of decision-making and data analysis nontrivial. As a result this area has attracted a lot of research interest with the aim being to yield accurate and time efficient and sensitive missing data imputation techniques especially when time sensitive applications are concerned like power plants and winding processes. In this article, considering arbitrary and monotone missing data patterns, we hypothesize that the use of deep neural networks built using autoencoders and denoising autoencoders in conjunction with genetic algorithms, swarm intelligence and maximum likelihood estimator methods as novel data imputation techniques  will lead to better imputed values than existing techniques. Also considered are the missing at random, missing completely at random and missing not at random missing data mechanisms. We also intend to use fuzzy logic in tandem with deep neural networks to perform the missing data imputation tasks, as well as different building blocks for the deep neural networks like Stacked Restricted Boltzmann Machines and Deep Belief Networks to test our hypothesis. The motivation behind this article is the need for missing data imputation techniques that lead to better imputed values than existing methods with higher accuracies and lower errors.
\end{abstract}


\section{Introduction}

Decision-making and data analysis tasks are made nontrivial by the presence of missing data in a database. The decisions made by decision makers are likely to be more accurate and reliable with complete datasets than with incomplete datasets containing missing data entries. Also, data analysis and data mining tasks yield more representative results and statistics when all the required data is available. As a result, there has been a lot of research interest in the domain of missing data imputation with researchers developing novel techniques to perform this task accurately and in a reasonable amount of time due to the time sensitive nature of some real life applications [\cite{aydilek2012novel}, \cite{rana2015robust}, \cite{koko2015missing}, \cite{mistry2009missing}, \cite{nelwamondo2007missing}, \cite{leke2014modeling}, \cite{mohamed2007estimating}, \cite{abdella2005use}, \cite{zhang2011missing} and \cite{zhang2011shell}]. Applications such as in medicine, manufacturing or energy that use sensors in instruments to report vital information that makes time sensitive decisions, may fail when there are missing data in the database. In such cases, it is very important to have a system capable of imputing the missing data from the failed sensors with high accuracy as quickly as possible. The imputation procedure in such cases requires the approximation of the missing value taking into account the interrelationships that exist between the values of other sensors in the system. There are several reasons that could lead to data being missing in a dataset. These could be as a result of data entry errors or respondents not answering certain questions in a survey during the data collection phase. Furthermore, failure in instruments and sensors could be a reason for missing data entries.  The table below depicts a database consisting of seven feature variables with the values of some of the variables missing. The variables are $X_{1}$, $X_{2}$, $X_{3}$, $X_{4}$, $X_{5}$, $X_{6}$ and $X_{7}$. 

\begin{table}[h]
	\caption{Database with Missing Data Entries}
	\label{tab:table1}
	\begin{center}
		\begin{tabular}{| c | c | c | c | c | c | c | c |}	
			\hline
			Instance & $X_{1}$ & $X_{2}$ & $X_{3}$ & $X_{4}$ & $X_{5}$ & $X_{6}$ & $X_{7}$ \\ \hline
			1 & 19 & ? & 14 & 49.8958 & 17.7759 & ? & 0.7717 \\ \hline
			2 & ? & 16 & 13 & ? & 23.7999 & 3.3254 & 0.2341 \\ \hline
			3 & 54 & 47 & ? & 55.8314 & ? & 12.6874 & 4.8522 \\ \hline
			4 & ? & 43 & 31 & 40.4672 & 18.4459 & 9.1189 & 3.0794 \\ \hline
			5 & 41 & ? & 27 & 18.0262 & 8.5707 & 0.4103 & 0 \\ \hline
			6 & 41 & 37 & 29 & 28.3564 & 6.9356 & 2.3057 & ? \\ \hline
			7 & 27 & 25 & 16 & 15.4484 & 9.1138 & ? & ? \\ \hline
			8 & 6 & 2 & 1 & 20.6796 & ? & ? & ? \\ \hline
			9 & 15 & 13 & 10 & ? & ? & ? & ? \\ 
			\hline
		\end{tabular}
	\end{center}
\end{table}

Consider that the database in question has several records of the seven variables with some of the data entries for some variables not available. The question of interest is, can we say with some degree of certainty what the missing data entries are? Furthermore, can we introduce techniques for approximation of the missing data when correlation and interrelationships between the variables in the database are considered? We aim to use deep learning techniques, Genetic Algorithms (GAs), Maximum Likelihood Estimator (MLE) and Swarm Intelligence (SI) techniques to approximate the missing data in databases with the different models created catering to the different missing data mechanisms and patterns. Therefore, with knowledge of the presence of interrelationships or lack thereof between feature variables in the respective datasets, one will know exactly what model is relevant to the imputation task at hand. Also we plan to use fuzzy logic with deep learning techniques to perform the imputation tasks.

\section{Background}
In this section, we present details on the problem of missing data and the Deep Learning techniques we aim to use to solve the problem.

\subsection{Missing Data} 
Missing data is a scenario in which some of the components of the dataset are not available for all feature variables, or may not even be defined within the problem domain in the sense that the values do not match the problem definition by either being outliers or inaccurate \citep{rubin1978multiple}. This produces a variety of problems in several application domains that rely on the access to complete and quality data. As a result, techniques aimed at handling the problem have been an area of research for a while in several disciplines [\cite{allison1999multiple}, \cite{little2014statistical} and \cite{rubin1978multiple}]. Missing data may occur in several ways in a dataset. For example, it may occur due to several participants’ non-response to questions in the data collection process or data entry process . There are also other situations in which missing data may occur due to failures of sensors or instruments in the data recording process for sectors that use these. The following subsubsections present the different missing data mechanisms.

\subsubsection{Missing Data Mechanisms}
The way to handle missing data in a reasonable manner depends on how the data points go missing. According to \cite{little2014statistical}, there exist three missing data mechanisms. They are: Missing Completely at Random (\textbf{MCAR}), Missing at Random (\textbf{MAR}), and a Missing not at Random or Non-Ignorable case(\textbf{MNAR or NI}).

\subsubsection*{Missing Completely at Random}
MCAR scenario arises when the chances of there being a missing data entry for a feature variable is not dependent on the feature variable itself or on any of the other feature variables in the dataset \citep{leke2014modeling}. This implies that the missing value is independent of the feature variable being considered or the other feature variables within the dataset \citep{rubin1978multiple}. In Table \ref{tab:table1}, the nature of the missing value in $X_{2}$ for row 5 is said to be MCAR if the nature of this missing value does not depend on $X_{1}$, $X_{3}$, $X_{4}$, $X_{5}$, $X_{6}$ and $X_{7}$ and the variable $X_{2}$ itself.

\subsubsection*{Missing at Random}
MAR occurs if the chances of there being a missing value in a specific feature variable depends on all the other feature variables within the dataset, but not on the feature variable of interest \citep{leke2014modeling}. MAR means the value for the feature variable is missing, but conditional on some other feature variable observed in the dataset, although not on the feature variable of interest \citep{scheffer2002dealing}. In Table \ref{tab:table1}, the nature of the missing value in $X_{2}$ is said to be MAR if the missing nature of the value depends on $X_{1}$, $X_{3}$, $X_{4}$, $X_{5}$, $X_{6}$ and $X_{7}$ but not on $X_{2}$ itself.

\subsubsection*{Missing Not at Random or Non-Ignorable Case}
The third type of missing data mechanism is the non-ignorable case. The non-ignorable case occurs when the chances of there being a missing entry in variable, $X_{2}$ for example, is influenced by the value of the variable $X_{2}$ regardless of whether or not the other variables in the dataset are altered and modified [\cite{leke2014modeling}, \cite{allison1999multiple}]. In this case, the pattern of missing data is not random and it is impossible to predict this missing data using the rest of the variables in the dataset. Non-ignorable missing data is the most difficult to approximate and model than the other two missing data mechanisms \citep{rubin1978multiple}. In Table \ref{tab:table1} the nature of the missing value in $X_{2}$ is said to be non-ignorable if the missing value in $X_{2}$ depends on the variable itself and not on the other variables.

\subsection{Missing Data Patterns}
There are two main missing data patterns defined by \cite{little2014statistical}. These patterns are the arbitrary and monotone missing data patterns. In the arbitrary missing data pattern, missing observations may occur anywhere and the ordering of the variables is of no importance as in rows 1 to 5. In monotone missing patterns, the ordering of the variables is of importance and occurrence is not random. In this case, if we have a dataset with variables as in Table \ref{tab:table1}, it is said to be a monotone missing pattern if a variable $X_{j}$ is observed for a particular scenario, and this implies that all the previous variables $X_{k}$, where $k < j$, are also observed for that scenario \citep{little2014statistical}. Table \ref{tab:table1} shows an arbitrary missing data pattern from rows 1 to 5 and a monotone missing data pattern from rows 6 to 9. In Table \ref{tab:table1} the missing values are random and can happen at any point in the dataset from rows 1 to 5 while it can be seen that missing values have some common order in rows 6 to 9. This means that if the values for a variable $X_{j}$ are missing, so are the values for other variables $X_{i}$, where $i > j$.

\subsection{Deep Learning}
Deep Learning comprises of several algorithms in machine learning that make use of a cataract of nonlinear processing units organized into a number of layers that extract and transform features from the input data [\cite{deng2013recent}, \cite{deng2014deep}]. Each of the layers use the output from the previous layer as input and a supervised or unsupervised algorithm could be used in the training or building phase. With these come applications in supervised and unsupervised problems like classification and pattern analysis respectively. It is also based on the unsupervised learning of multiple levels of features or representations of the input data whereby higher-level features are obtained from lower level features to yield a hierarchical representation of the data \citep{deng2014deep}. By learning multiple levels of representations that depict different levels of abstraction of the data, we obtain a hierarchy of concepts.

\begin{figure}[h!]
	\includegraphics[width = \textwidth]{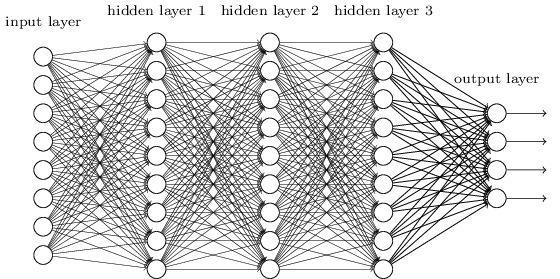}
	\caption{Deep Neural Network Architecture}
	\label{fig:figure1}
\end{figure}

\begin{figure}[h!]
	\includegraphics[width = \textwidth]{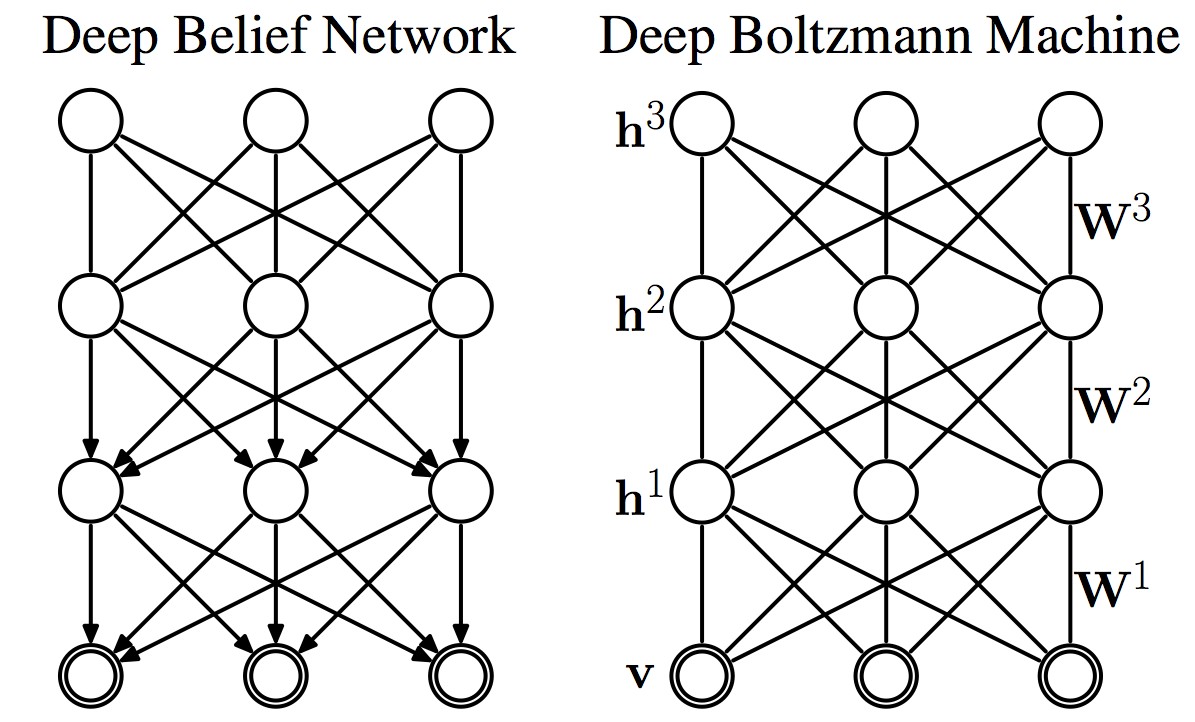}
	\caption{Deep Belief Network (Left) and Deep/Stacked Restricted Boltzmann Machine (Right) Architectures where W are weights and h denotes hidden layers [\cite{salakhutdinov2009deep}]}
	\label{fig:figure2}
\end{figure}

\begin{figure}[b]
	\includegraphics[width = \textwidth]{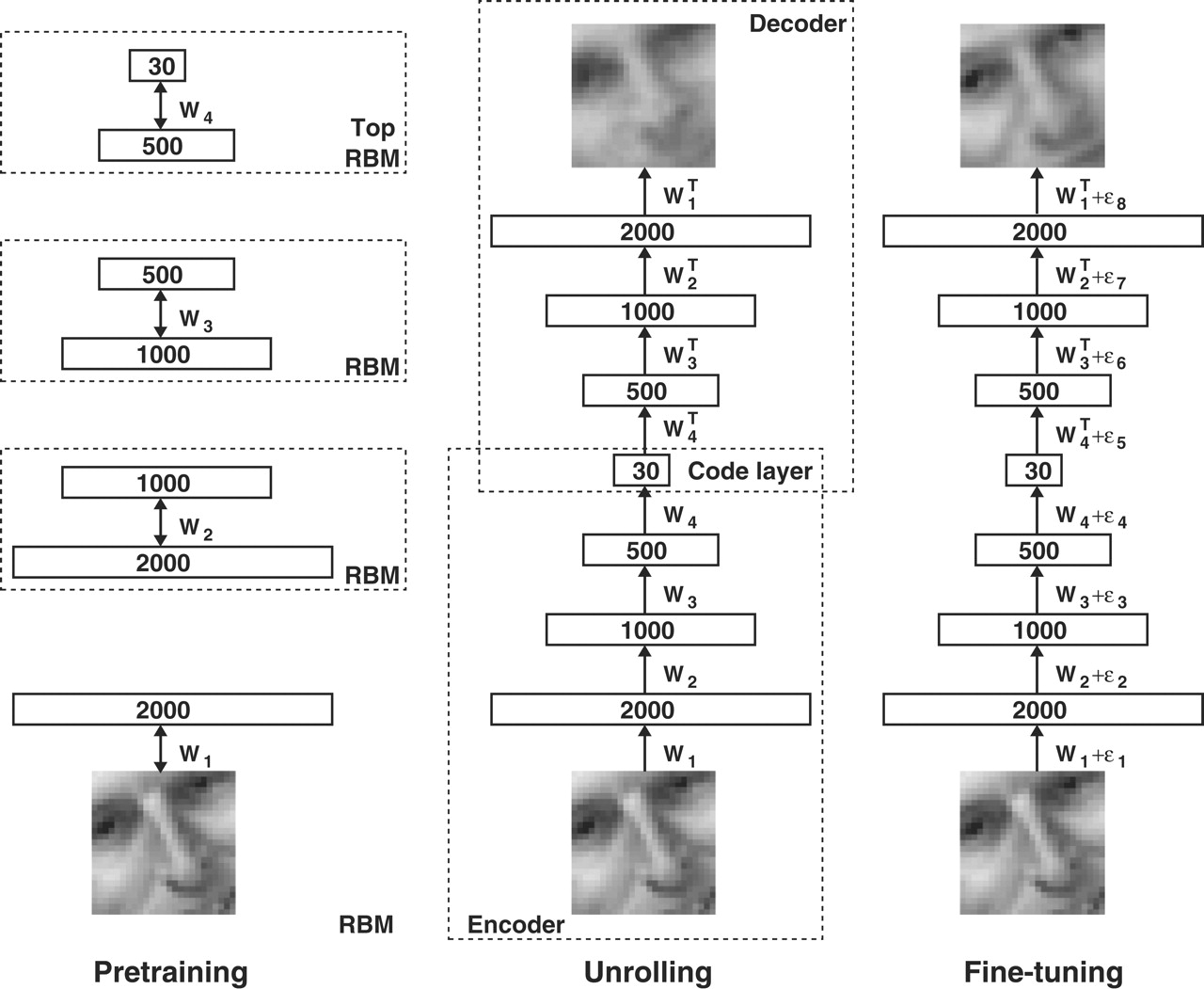}
	\caption{Individual Restricted Boltzmann Machines (Left), Stacked Restricted Boltzmann Machines forming Encoder and Decoder (Center) and Trained Auto-Encoder Deep Neural Network (Right) [\cite{hinton2006reducing}]}
	\label{fig:figure3}
\end{figure}

There are different types of Deep Learning architectures such as Convolutional Neural Networks (CNN), Convolutional Deep Belief Networks (CDBN), Deep Neural Networks (DNN), Deep Belief Networks (DBN), Stacked (Denoising) Auto-Encoders (SAE/SDAE) and Deep/Stacked Restricted Boltzmann Machines (DBM). We intend to make use of DNNs and SAEs predominantly, and the others with the exception of CNNs and CDBNs. DNNs are commonly understood in terms of the Universal Approximation Theorem, Probabilistic Inference or Discrete Signal Processing. An artificial neural network (ANN) with numerous hidden layers of nodes between the input layer and the output layer is known as a DNN. They are typically designed as feed forward networks and can be trained discriminatively utilizing standard back propagation with updates of the weights being done by use of stochastic gradient descent. Typical choices for the activation and cost functions are the softmax and cross entropy functions for classification tasks, with sigmoid and standard error functions used for regression or prediction tasks with normalized inputs. In Figures \ref{fig:figure1}-\ref{fig:figure3}, the architectures of four deep learning techniques are depicted. Figure \ref{fig:figure1} shows a DNN with eight input nodes in the input layer, three hidden layers each with nine nodes and an output layer with four nodes. The nodes from each layer are connected with those from the subsequent and preceding layers. Figure \ref{fig:figure2} shows a DBN and a DBM whereby the first layer of nodes (bottom-up) is the input layer (v) with visible units representing the database feature variables and the subsequent layers of nodes are binary hidden nodes (h). The arrows in the DBN indicate that the training is a top-down approach while the lack of arrows in the DBM is a result of the training being both top-down and bottom-up. In Figure \ref{fig:figure3}, we see individual RBMs being stacked together to form the encoder part of an autoencoder, which is transposed to yield the decoder part. The autoencoder is then fine-tuned using back propagation to modify the interconnecting weights with the aim being to minimize the network error.  

\subsection{Related Work}
In this section, we present some of the work that has been done by researchers to address the problem of missing data. In \cite{zhang2011missing}, it is suggested that information within incomplete cases, that is, instances with missing values be used when estimating missing values. A nonparametric iterative imputation algorithm (NIIA) is proposed that leads to a root mean squared error value of at least 0.5 on the imputation of continuous values and a classification accuracy of at most 87.3\% on the imputation of discrete values with varying ratios of missingness. \cite{lobato2015multi} present a multi-objective genetic algorithm approach for missing data imputation. It is observed that the results obtained outperform some of the well known missing data methods with accuracies in the 90 percentile. In \cite{zhang2011shell}, the shell-neighbor method is applied in missing data imputation by means of the Shell-Neighbor Imputation (SNI) algorithm which is observed to perform better than the k-Nearest Neighbor imputation method in terms of imputation and classification accuracy as it takes into account the left and right nearest neighbors of the missing data as well as varying number of nearest neighbors contrary to k-NN that considers just fixed k nearest neighbors. \cite{rana2015robust} use robust regression imputation for missing data in the presence of outliers and investigate its effectiveness. \cite{abdella2005use} implement a hybrid genetic algorithm-neural network system to perform missing data imputation tasks with varying number of missing values within a single instance while \cite{aydilek2012novel} create a hybrid k-Nearest Neighbor-Neural Network system for the same purpose. In some cases, neural networks were used with Principal Component Analysis (PCA) and genetic algorithm as in \cite{mistry2009missing}, \cite{mohamed2007estimating} and \cite{nelwamondo2007missing}. \cite{leke2014modeling} use a hybrid of Auto-Associative neural networks or autoencoders with genetic algorithm, simulated annealing and particle swarm optimization to impute missing data with high levels of accuracy in cases where just one feature variable has missing input entries. Novel algorithms for missing data imputation and comparisons between existing techniques can be found in papers such as \cite{schafer2002missing}, \cite{liew2011missing}, \cite{myers2011goodbye}, \cite{lee2010multiple}, \cite{baraldi2010introduction}, \cite{van2012flexible}, \cite{jerez2010missing} and \cite{kalaycioglu2015comparison}.

\section{Theoretical Model}
\begin{figure}[h]
	\includegraphics[width = \textwidth]{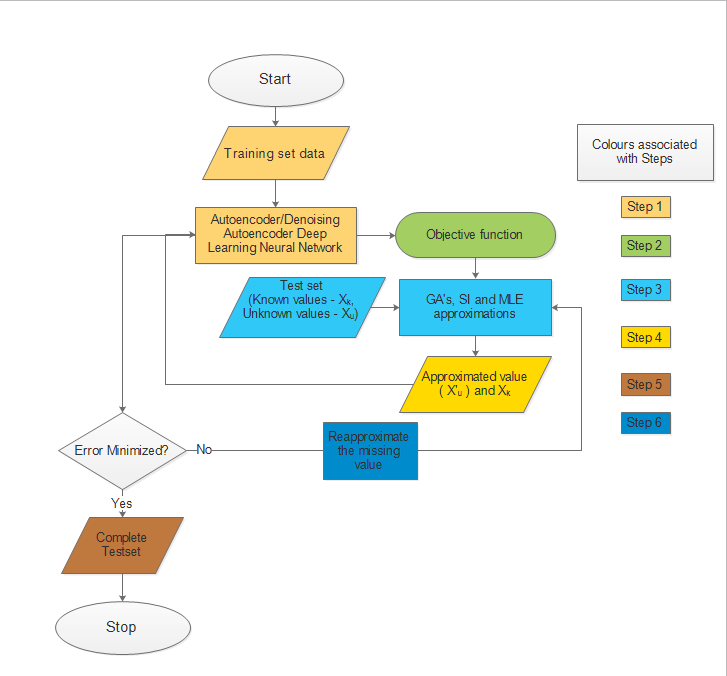}
	\caption{Flowchart Representation of Theoretical Model}
	\label{fig:figure4}
\end{figure}

In this section, we outline the methodology used to address the problem  of missing data. The approach used to design the novel imputation techniques with SAE/SDAE involves the following six steps which are depicted in figure \ref{fig:figure4}:

\begin{enumerate}
	\item Train the Deep Neural Network with a complete set of records to recall the inputs as the outputs. Inputs are the dataset feature variables, for example $X_{1}$ to $X_{7}$ in Table \ref{tab:table1}, and the outputs are these same feature variables as the aim is to reproduce these inputs at the output layer. For the network to be able to do this, it needs to extract information from the input data, which is captured in the updated network weights and biases. The extraction of information is done during the training phase whereby lower level features are extracted from the input data after which low-level features are extracted till high-level features are obtained yielding a hierarchical representation of the input data. The overall idea is that features are extracted from features to get as good a representation of the data as possible. In the encoder phase mentioned in the previous section, a deterministic mapping function, $f_{\theta}$, creates a hidden representation, $y$, of the input data $x$. It is typically represented by an affine mapping and subsequently a nonlinearity, $f_{\theta} \left(x\right) = s\left(Wx + b\right)$ \citep{isaacs2014rep}. The $\theta$ parameter comprises of the matrix of weights $W$ and the vector of offsets/biases $b$. In the decoder phase, $y$ being the hidden representation is remapped to $z$ which is a vector reconstruction in the input space with $z = g_{\theta^{'}}\left(y\right)$ \citep{isaacs2014rep}. The function $g_{\theta^{'}}$ is the decoder function which is an affine mapping deliberately ensued by a non-linearity with squashing traits that either follows the form $g_{\theta^{'}}\left(y\right) = W'y + b'$ or $g_{\theta^{'}}\left(y\right) = s\left(W'y + b'\right)$ with the parameter set $\theta^{'}$ comprising of the transpose of the weights and biases from the encoder \citep{isaacs2014rep}.
	\item Obtain the objective function from step 1 as depicted in Figure \ref{fig:figure4} as input to the optimization techniques. The updated weights and biases mentioned in step 1 are gotten by back propagating the error at the output layer obtained by comparing the actual output to the network output through the network. The function or equation used to compare the actual output to the network output is used as the objective function. $z$ from step 1 is not explained as a rigorous regeneration of $x$ but rather as the parameters of a distribution $p\left(X|Z = z\right)$ in probabilistic terms, that may yield $x$ with high probability \citep{isaacs2014rep}. This thus leads to $p\left(X|Y = y\right) = p\left(X|Z = g_{\theta^{'}}\left(y\right)\right)$. From this, we obtain an associated reconstruction error which is to be optimized by the optimization techniques and is of the form $L\left(x, z\right) \propto -logp\left(x|z\right)$. This equation could also be written as $\delta_{AE}\left(\theta\right) = \Sigma_{t}L\left(x^{\left(t\right)},g_{\theta}\left(f_{\theta}\left(x^{\left(t\right)}\right)\right)\right)$ \citep{bengio2013representation}. For a denoising autoencoder, the reconstruction error to be optimized is expressed as $\delta_{DAE} = \Sigma_{t} \mathbb{E}_{q\left(\tilde{x}|x^{\left(t\right)}\right)}\left[ L\left(x^{\left(t\right)},g_{\theta}\left(f_{\theta}\left(\tilde{x}\right)\right)\right)\right]$ where $\mathbb{E}_{q\left(\tilde{x}|x^{\left(t\right)}\right)}\left[.\right]$ averages over the corrupted examples $\tilde{x}$ drawn from a corruption process $q\left(\tilde{x}|x^{\left(t\right)}\right)$ \citep{bengio2013representation}.
	\item Approximate the missing data entries using the approximation techniques. MCAR, MAR and MNAR missing data mechanisms will be considered as well as arbitrary and monotone missing data patterns. Different models will be created to experiment with these and test the hypothesis. In testing the hypothesis, we use the test set of data which consist of known feature variable values $X_{k}$ and unknown or missing feature variable values $X_{u}$ as input to the trained deep learning technique. The $X_{k}$ values are passed as input to the network while the $X_{u}$ values are first estimated by the approximation techniques before being passed into the network as input. The optimal $X_{u}$ value is obtained when the objective function from step 2 is minimized. 
	\item Use the now completed database with the approximated missing values in the trained Deep Learning method from step 1 to observe whether or not the objective has been minimized. In this case, that will be checking if the error is minimized as we attempt to reconstruct the input. 
	\item If so, the complete dataset is presented as output.
	\item If not, do step 3. 
\end{enumerate}

\section{Possible Benefits}
In this research, we are introducing novel data imputation techniques which we expect will be of benefit to the research community interested in missing data imputation. Some of these are:
\begin{itemize}
	\item With the techniques introduced, we expect to yield improved missing data imputation accuracies compared against existing methods by looking at the relative prediction accuracy, correlation coefficient, standard square error, mean and root mean squared errors and other relevant representative metrics in comparison to existing techniques. This expectation stems from the manner in which deep learning methods extract information and features from the input data.
	\item With literature stating that deep neural networks are capable of representing and approximating more complex functions and relations than simple neural networks, we hope these techniques will be applicable in a variety of sectors regardless of the complexity of the problem with high accuracy. This will be tested against existing techniques and the aforementioned.
	\item Possible parallelization of the imputation tasks using the methods to be introduced could lead to faster imputed missing values which benefits time sensitive applications.
\end{itemize}

\section{Possible Limitations}
Although there are possible benefits to using the novel techniques to be introduced, there could possibly be limitations observed, for example:
\begin{itemize}
	\item Using Deep Neural Networks could possibly lead to a lot of time being required to do the imputations and obtaining a complete dataset due to the number of parameters that need to be optimized during training and also the number of computations done during testing. The full effect of long computation times could be felt in time sensitive applications such as in medicine, finance or manufacturing. The slow computation time could be addressed by parallelizing the processes on a multicore system. Each core could handle the imputation of the missing data value(s) in different rows depending on the number of cores. Also, dynamic programming could be used to speed up the computation time.
	\item Besides time being a factor, there could also be a problem of space required to do the computations. To address these two drawbacks, a complexity analysis will be done to verify the time and space complexities of the proposed methods. Anything less than $O\left(n^{2}\right)$ will be preferable with $O\left(n\log n\right)$ being regarded as the ideal complexity for both.
\end{itemize}

\section{Conclusion}
In this article, we propose a new hypothesis that the use of deep learning techniques in conjunction with swarm intelligence, genetic algorithms and maximum likelihood estimator methods will lead to better imputations due to the fact that a hierarchical representation of the input data is obtained as higher level features are further extracted from lower level features in deep learning methods. This hypothesis is investigated by taking into account a comparison between the techniques to be introduced and the existing methods like Neural Networks with Genetic Algorithm, Auto-Associative Neural Network with Genetic Algorithm, K-Nearest Neighbor with Neural Networks, Neural Networks with Principal Component Analysis and Genetic algorithm and so on. The main motivation behind this hypothesis is the need to provide datasets with highly representative and accurate feature values from which trustworthy decisions and data analytics and statistics will emerge.

\textbf{}
\newpage

\vskip 0.2in

\end{document}